\documentclass{article}

\usepackage{PRIMEarxiv}

\usepackage[utf8]{inputenc} % allow utf-8 input
\usepackage[T1]{fontenc}    % use 8-bit T1 fonts
\usepackage{hyperref}       % hyperlinks
\usepackage{url}            % simple URL typesetting
\usepackage{booktabs}       % professional-quality tables
\usepackage{amsfonts}       % blackboard math symbols
\usepackage{nicefrac}       % compact symbols for 1/2, etc.
\usepackage{microtype}      % microtypography
\usepackage{lipsum}
\usepackage{fancyhdr}       % header
\usepackage{graphicx}       % graphics
\graphicspath{{media/}}     % organize your images and other figures under media/ folder

\title{Reducing a complex two-sided smartwatch examination for Parkinson's Disease to an efficient one-sided examination preserving machine learning accuracy}

\author{
  Alexander Brenner$^a$, Michael Fujarski$^a$, Tobias Warnecke$^b$, Julian Varghese$^a$ \\
  $^a$Institute of Medical Informatics, University of Münster, Münster, Germany\\
  $^b$Department of Neurology and Neurorehabilitation, Klinikum Osnabrück – Academic
teaching hospital of \\ the University of Münster, Osnabrück, Germany
}

\begin{document}
\maketitle

\begin{abstract}
Sensors from smart consumer devices have demonstrated high potential to serve as digital biomarkers in the identification of movement disorders in recent years. With the usage of broadly available smartwatches we have recorded participants performing technology-based assessments in a prospective study to research Parkinson's Disease (PD). In total, 504 participants, including PD patients, differential diagnoses (DD) and healthy controls (HC), were captured with a comprehensive system utilizing two smartwatches and two smartphones. To the best of our knowledge, this study provided the largest PD sample size of two-hand synchronous smartwatch measurements. To establish a future easy-to use home-based assessment system in PD screening, we systematically evaluated the performance of the system based on a significantly reduced set of assessments with only one-sided measures and assessed, whether we can maintain classification accuracy. 
\end{abstract}


\section{Introduction}
Consumer grade smart devices are broadly available and became attractive for various classification tasks, e.g. human activity recognition via wearable sensors, smartphones or cameras \cite{straczkiewicz2021systematic}. Utilizing a smart device system (SDS) consisting of consumer devices, we have conducted a prospective study to research Parkinson‘s Disease (PD), which is one of the most common neurological disorders. Among various non-motor symptoms, PD affects the patient's movement with symptoms such as slowed movement, rigidity and tremor. Along with other movement disorders, PD is primarily diagnosed on the basis of clinical examination and nuclear imaging. The research goal of the SDS study has been the exploration of digital biomarkers using sensor data and the evaluation of their diagnostic potential. In the study, a total of 504 participants, including PD patients, differential diagnosis (DD) and healthy controls (HC), have been recorded from 2018 to 2021. Each recording session consists of a series of measures from different movement tasks designed by movement disorder experts. Each task was synchronously measured with two smartwatches, one attached to each hand. To the best of our knowledge, the dataset provides the largest PD sample size of two-hand smartwatch measurements. Varghese et al. have already validated sensor quality of the utilized smartwatches in comparison to high-precision seismometers and demonstrated their potential in classifying PD \cite{varghese2021sensor}. For the utilization as simple screening devices, further validation of simplicity, reliability and interpretability of the system is needed. This work therefore focuses on the systematic evaluation based on a reduced set of different movement records and one-sided measures to assess importance and reduce redundancy. We hypothesize that two smartwatches potentially increase knowledge in classification as PD is often observed with side-dominant symptoms. In this work we investigate this potential knowledge gain and evaluate the diagnostic potential of smartwatch-based measurements. Since a simplified set of sensors reduces costs and complexity of the overall system, it is important to evaluate the possible accuracy loss of a more simple system. Thus, our research objective is to derive digital biomarkers from the sensor measures to reliably predict movement disorders and to elaborate on whether a one-sided setup is sufficient for this task.

\section{Methods}
\subsection{Data}
The study has been registered (ClinicalTrials.gov ID: NCT03638479) and approved by the ethical board of the University of Münster and the physician’s chamber of Westphalia-Lippe (Reference number: 2018-328-f-S). The study records consists of three patient groups: 1) Parkinson’s disease (PD), including a broad range of different PD progress states according to Hoehn and Yahr \cite{bhidayasiri2012parkinson}, 2) differential diagnoses (DD) and 3) healthy controls (HC). All diagnoses were confirmed by neurologists and reviewed by a senior movement disorder expert. In addition to the smartwatch measures, each participant filled out a basic questionnaire with information about age, medical history, handedness and more. Since we focus on the sensor measures derived from the smartwatches, we only consider information about handedness from the questionnaire data. Around 8\% of all study participants stated that they are left-handed. The overview of the data sample is presented in \autoref{tab:1}.

\begin{table}[h]
\caption{Participant sample.}
\label{tab:1}
\centering
\begin{tabular}{c|c|c}
\hline
Group                       & \#Right-handed & \#Left-handed \\ \hline
Parkinson's Disease (PD)    & 262            & 17            \\
Differential diangosis (DD) & 122            & 12            \\
Healthy control (HC)        & 80             & 11            \\ \hline
All                         & 464            & 40            \\ \hline
\end{tabular}
\end{table}

\subsection{Feature computation and Machine Learning}
We implemented a Machine Learning (ML) pipeline to evaluate diagnostic potential of the sensor data. Therefore, we performed a pre-processing procedure to organize the data for classification. In total, 11 assessment steps were performed from which 3 took 20 seconds and the rest 10 seconds. Details of the tasks can be derived from the original study design from Varghese et al. \cite{varghese2019smart}. The 20 seconds long records were cut into two parts so that we achieved 14 time series of 10 seconds length per participant. Each of these 14 time series were recorded synchronously with one smartwatch attached to each wrist respectively. The smartwatches recorded acceleration and rotation data. Each sensor stored a time series over three spatial axes (x, y, z). With this setup we achieved 168 channels of time series data for each participant (14 tasks * 2 arms * 2 sensor * 3 axes = 168). For each channel representative features were computed to form the input to the ML algorithms. Feature extraction was composed of the following steps: 1) On each time series the power spectral density (PSD) was computed in discrete 1 Hertz steps using Welch’s method. Values for 0 Hertz or above 19 Hertz were discarded, resulting in 19 frequency features per channel that were scaled by logarithm afterwards. 2) Each time series was split into 4 parts of equal length. For each of the segments the standard deviation, the maximum absolute amplitude and the sum of absolute energy was computed, resulting in 12 features per channel. In conclusion, each channel was transformed into a feature vector with 31 elements, so that each sample was represented with a feature vector of 5208 features (168 channels * 31 features). The resulting feature vector was normalized to unit variance afterwards.

For classification we used the scikit-learn implementation of the support-vector machine (SVM) \cite{sklearn_api}. To obtain representative results, a 3 times randomly repeated 5-fold cross validation was performed for every classification. Further, hyperparameters of the SVM were selected via grid-search for every changing input setting, the parameter grid was defined with: 'C': [0.1, 1.0, 10.0, 100.0, 1000], 'gamma': [0.000001, 0.00001, 0.0001, 'scale', 0.001]. Four relevant classification tasks were cross-validated: 1) PD vs. HC, 2) Movement disorders (PD + DD) vs. HC, 3) PD vs. DD and 4) a multi-class setting (PD vs. DD vs. HC). Since the number of samples in different target groups were imbalanced, class weights were balanced in the training process and classification performance was compared based on balanced accuracy.

\subsection{Feature selection}
The baseline performances for the four classification tasks were established using the complete set of sensor measures. To assess whether a single-sided setup using only one smartwatch reduces performance, 5 settings were tested: a) Using input from both arms, b) from the left arm only, c) from the right arm only, d) from the strong arm and e) from the weak arm. The strong arm corresponds to the stated handedness in the questionnaire. The balanced classification accuracy for each setting was then compared to the baseline score.
\newpage

\begin{figure}[ht!]
	\centering
		\includegraphics[width=0.65\textwidth]{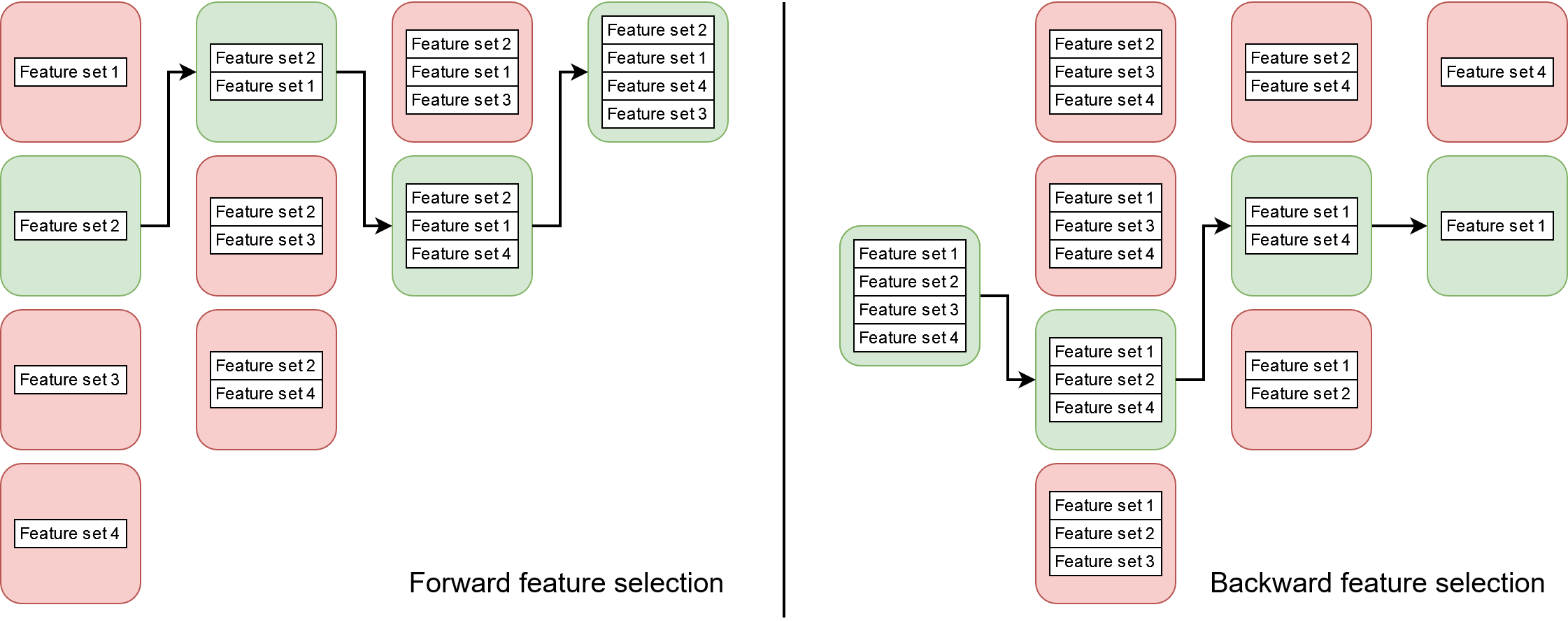}
	\caption{Visualization of the forward and backward feature selection procedure. Features are iteratively added/removed by comparing balanced accuracy and selecting the best performing set of features in each step.}
	\label{fig:0}
\end{figure}

To evaluate how the system potentially can be simplified in terms of reducing the set of tasks, we performed a systematic feature selection process. This process was composed of a forward and a backward feature selection procedure that is visualized in \autoref{fig:0}. We defined a feature group composed of all computed features that were derived from a certain movement task. Respectively, we divided the ML input into 11 feature groups. In the forward feature selection the best performing feature group was choosen. In the next iterations the feature group that best increased the classification score was iteratively added to the input set. As a result, we achieved 11 classification scores for consecutively larger sets of performed movement tasks. The backward feature selection worked similar, but instead of adding feature groups, we started with the complete set of feature groups and iteratively removed the feature group with least information gain.

To combine the insights from both analyses described above, the task reduction was further repeated using the best performing arm. Additionally, the best performing arm was selected on a reduced set of tasks with the above procedure for each classification setup respectively.

Given the resulting classification scores per feature-group we selected the smallest subsets that already achieved the best score ($\pm$ 0.5\%). Per classification task the selected sets were afterwards compared to find assessment steps that were most often excluded and the ones that were most often included.

\section{Results}
The baseline results for the complete set of input features are summarized in \autoref{tab:2}.

\begin{figure}[ht!]
	\centering
		\includegraphics[width=\textwidth]{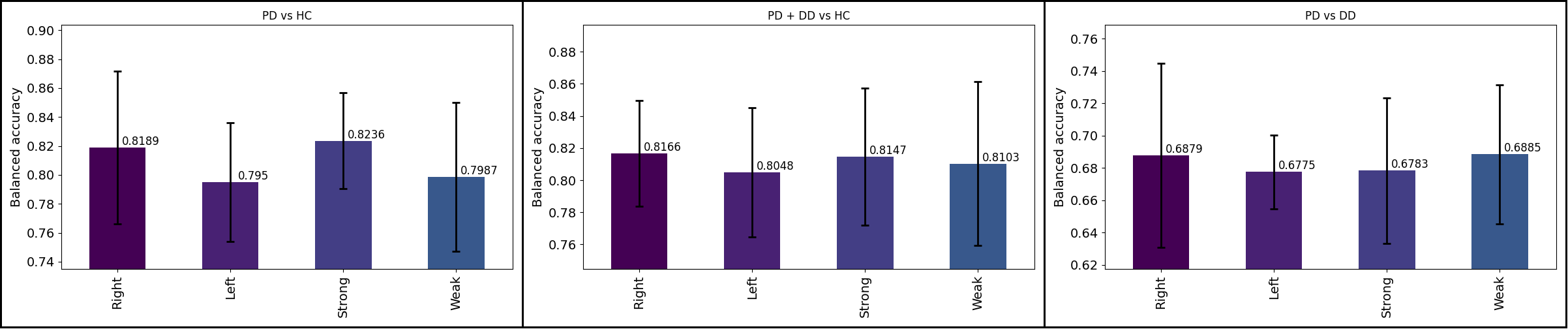}
	\caption{Results on four classification tasks using only one measured arm as input: 1) PD vs. HC, 2) Movement disorders (PD + DD) vs. HC, 3) PD vs. DD and 4) a multi-class setting (PD vs. DD vs. HC). Bar height indicate mean balanced accuracy, error bars represent the standard deviation.}
	\label{fig:1}
\end{figure}

The tests for different recording arms are displayed in \autoref{fig:1}. While for each classification task a different setting performed best, performance for the right arm only outperformed the left arm in every test case. The right-arm only setup is compared to the two-sided baseline in \autoref{tab:2}.

\begin{figure}[ht!]
	\centering
		\includegraphics[width=\textwidth]{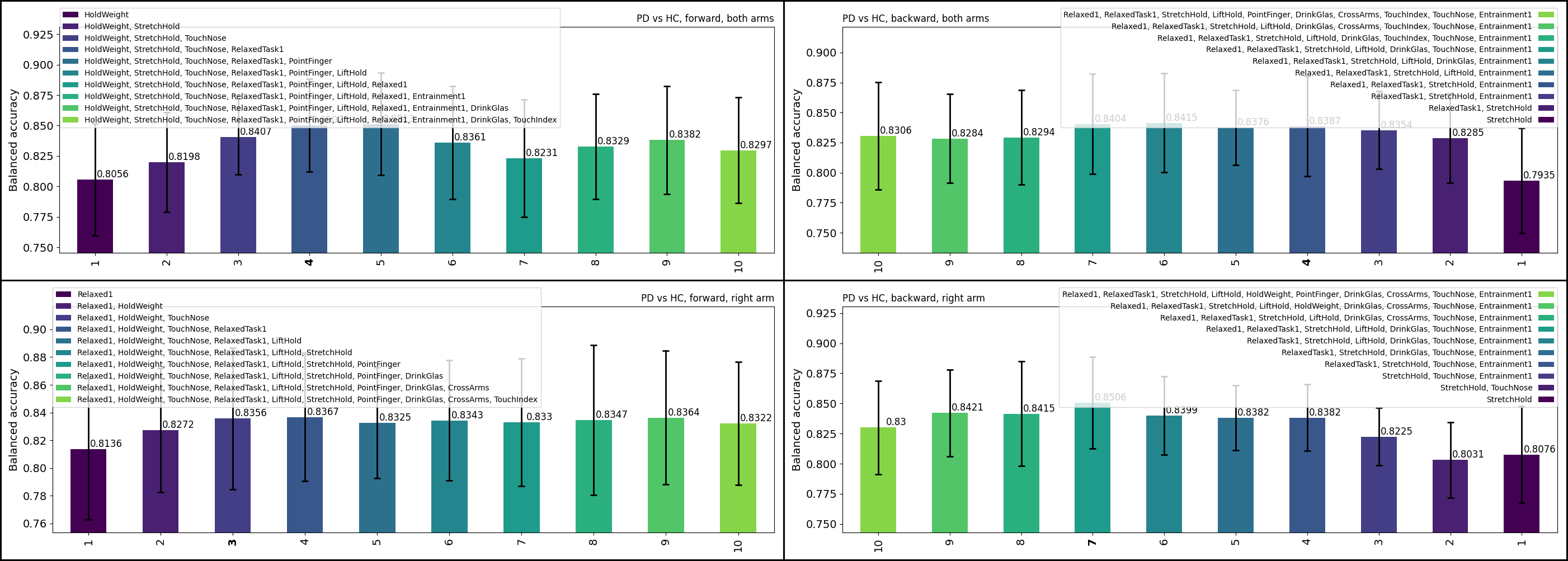}
	\caption{Results of forward and backward feature selection for the classification task: PD vs. HC. Bar height indicate mean balanced accuracy, error bars represent the standard deviation.}
	\label{fig:2}
\end{figure}

\begin{figure}[ht!]
	\centering
		\includegraphics[width=\textwidth]{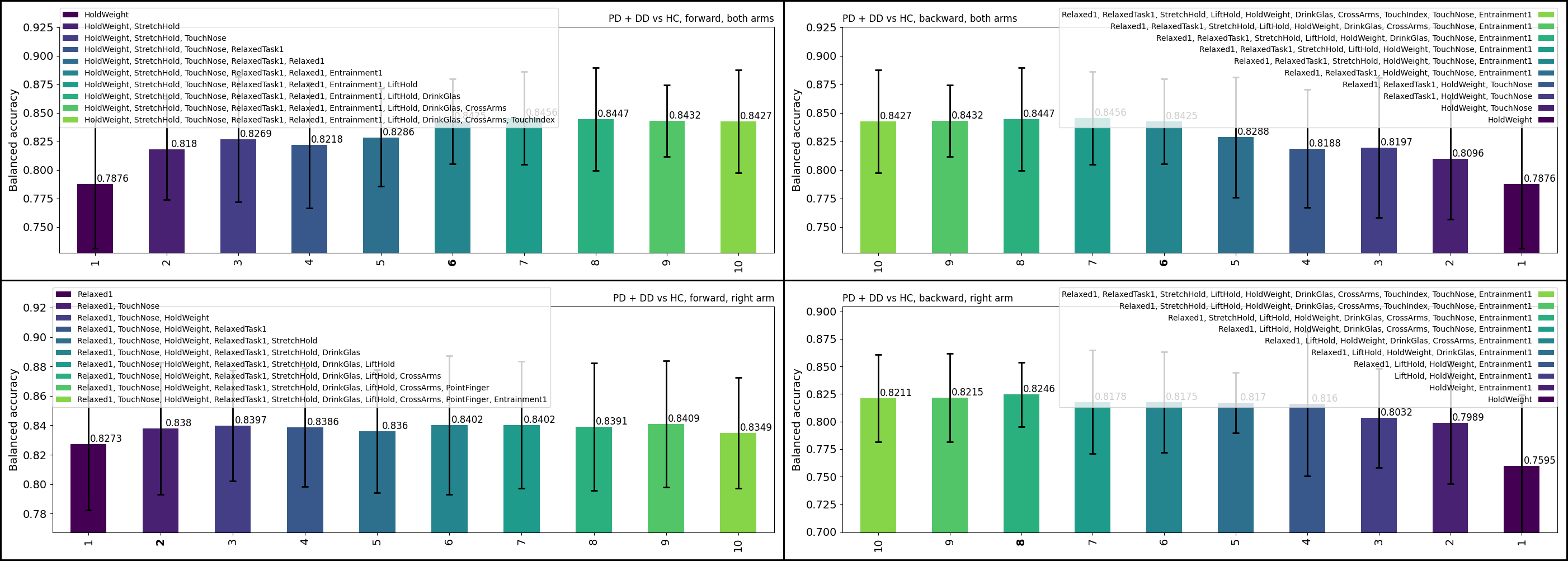}
	\caption{Results of forward and backward feature selection for the classification task: PD + DD vs. HC. Bar height indicate mean balanced accuracy, error bars represent the standard deviation.}
	\label{fig:3}
\end{figure}

\begin{figure}[ht!]
	\centering
		\includegraphics[width=\textwidth]{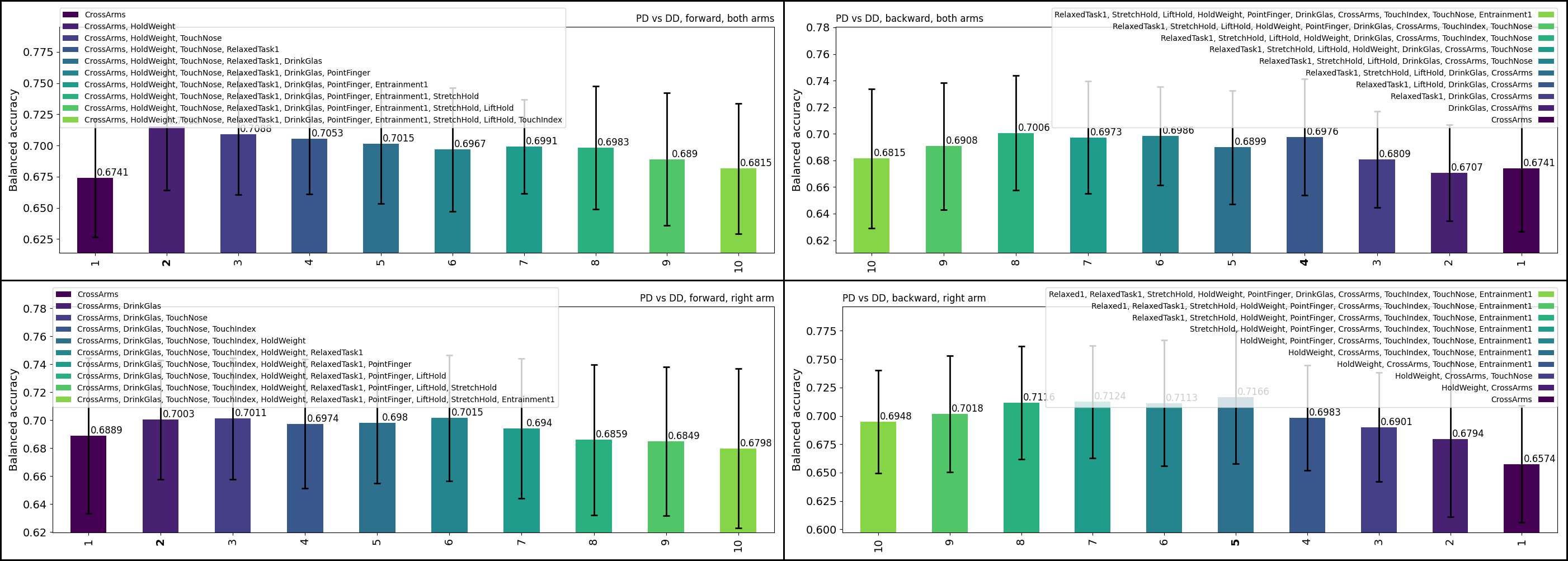}
	\caption{Results of forward and backward feature selection for the classification task: PD vs. DD. Bar height indicate mean balanced accuracy, error bars represent the standard deviation.}
	\label{fig:4}
\end{figure}

The forward and backward feature selection results for movement tasks are summarized in \autoref{fig:2} for PD vs. HC, \autoref{fig:3} for PD + DD vs. HC and \autoref{fig:4} for PD vs. DD. Feature selection was performed for including both arms and the right arm only.

\begin{figure}[ht!]
	\centering
		\includegraphics[width=\textwidth]{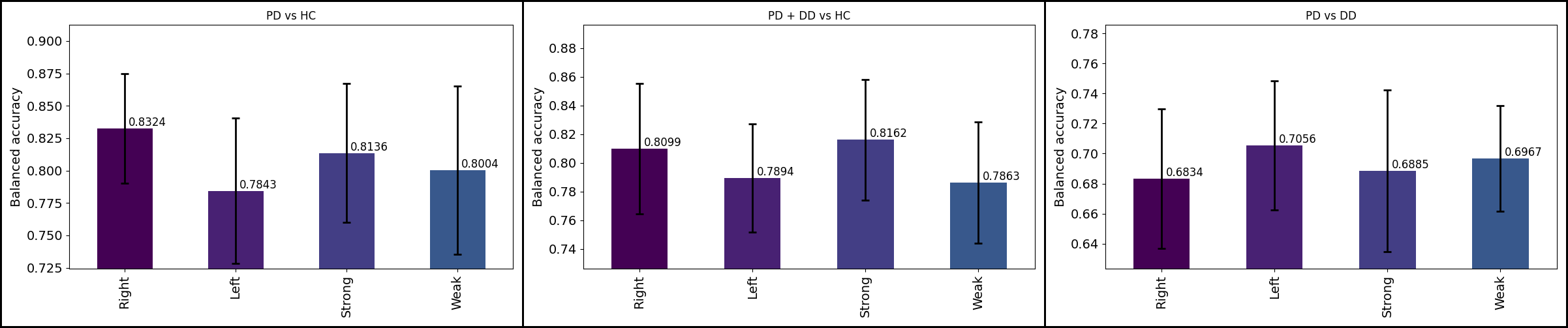}
	\caption{Results on four classification tasks using a reduced set of assessment steps and only one measured arm as input: 1) PD vs. HC, 2) Movement disorders (PD + DD) vs. HC, 3) PD vs. DD and 4) a multi-class setting (PD vs. DD vs. HC). Bar height indicate mean balanced accuracy, error bars represent the standard deviation.}
	\label{fig:5}
\end{figure}

Three assessment steps were frequently excluded in the selection of the best subset, consistently over all classification tasks. With removing these from the input data, the recording site experiments were repeated. Results are shown in \autoref{fig:5}. In two out of three classification tasks the right-side setup outperformed the left-side setup.

\begin{table}[ht!]
\caption{Performance for the tasks (1) PD vs. HC, (2) Movement disorders (PD + DD) vs. HC and (3) PD vs. DD. All results are based on SVM classifier optimized via gridsearch. Values correspond to mean balanced accuracy (SD).}
\centering
\begin{tabular}{c|c|c|c|c}
\hline
               & Baseline        & Reduced task set & Right arm only  & Reduced task set, right arm only \\ \hline
PD vs. HC      & 0.8268 (0.0373) & 0.8225 (0.0471)  & 0.8189 (0.0528) & 0.8324 (0.0424)                  \\ \hline
PD + DD vs. HC & 0.8265 (0.0375) & 0.8264 (0.0495)  & 0.8166 (0.0328) & 0.8099 (0.0455)                  \\ \hline
PD vs. DD      & 0.6765 (0.0492) & 0.6895 (0.0506)  & 0.6879 (0.0569) & 0.6834 (0.0465)                  \\ \hline
\end{tabular}
\label{tab:2}
\end{table}

\section{Discussion}
We have performed a forward and backward feature selection grouping features by assessment steps and recording arm. Balanced classification performance was set as measure of information gain for the feature groups. The step-wise optimization has shown that a reduced set of movements achieved similar performance compared to the baseline. While we found differences in the importance of certain movements depending on the classification task, we still observed a subset of relevant features for all classification tasks. Three assessment steps were consistently excluded from the minimal feature sets that still achieved best classification performance given a tolerance of 0.5\%. These assessment steps were selected by comparing forward and backward feature selection given data from two arms and the right arm only.

Reducing the hardware to a single smartwatch generally showed only a marginal effect on classification accuracy. Performances were compared over different classification tasks selecting data from the right side, the left side or by handedness. With both, the complete and the reduced set of assessment steps, there was no clear outperforming selection of recording side. Still, we observed that the right-side outperformed the left-side in most cases. Hence, the right-sided setup was used in the final comparison to the baseline.

The summary of the results in \autoref{tab:2} shows that we successfully reduced the set of features while maintaining classification performance. For the classification task PD vs. HC balanced classification accuracy slightly increased from 82.68\% to 83.24\% when comparing the baseline result with the simplified version. Similarly, performance for PD vs. DD increased from 67.65\% to 68.34\% using the new simplified setup. Only for Movement Disorders vs. HC performance slightly dropped from 82.65\% to 80.99\%. While all performance differences are below 2\%, which is close to the standard deviation, the results still indicate the robustness of the reduced setup.

\section{Conclusion}
Based on our analysis we identified a reduced assessment setting compared to our original study. This will reduce the time of active assessment while maintaining accuracy. Further, we have validated one-sided smartwatch measures as a valuable option for classifying PD. These changes reduce complexity of the SDS, making it more practical for routine-screening and potential home-based assessments in the future.
\newpage

%Bibliography
\bibliographystyle{unsrt}  
\bibliography{references}

\end{document}